\documentclass[american,english]{article}
\usepackage[T1]{fontenc}
\usepackage[latin9]{inputenc}
\usepackage{color}
\usepackage{babel}
\usepackage{array}
\usepackage{float}
\usepackage{multirow}
\usepackage{algorithm2e}
\usepackage{varwidth}
\usepackage{amsmath}
\usepackage{amssymb}
\usepackage{graphicx}
\usepackage{geometry}
\PassOptionsToPackage{normalem}{ulem}
\usepackage{ulem}
\usepackage[pdfusetitle,
 bookmarks=false,
 breaklinks=false,pdfborder={0 0 1},backref=false,colorlinks=true]
 {hyperref}

\makeatletter

\providecommand{\tabularnewline}{\\}

\usepackage{cite}
\usepackage{amsmath,amssymb,amsfonts}
\usepackage{algorithmic}
\usepackage{graphicx}
\usepackage{textcomp}
\usepackage{xcolor}
\usepackage{balance}

\usepackage{colortbl} 
\definecolor{header_color}{rgb}{0.74,0.88,0.91}
\definecolor{even_color}{rgb}{0.9,0.9,0.9}
\definecolor{subheader_color}{rgb}{0.85,0.93,0.95}
\definecolor{childheader_color}{rgb}{1.0,0.93,0.87}

\ifdefined\showcaptionsetup
 \PassOptionsToPackage{caption=false}{subfig}
\fi
\usepackage{subfig}
\makeatother

\begin{document}
\title{SMOTE-VAR: An Uncertainty-Aware Oversampling Method for Predicting
Depression Remission in University Students}
\author{Dang Nguyen\textsuperscript{1}\thanks{Corresponding author: Dang Nguyen (d.nguyen@deakin.edu.au)},
Arun Kumar A V\textsuperscript{1}, Taylor A. Braund\textsuperscript{2},
Wu Yi Zheng\textsuperscript{2}, Debopriyo Bal\textsuperscript{2},\\
Leonard Hoon\textsuperscript{1}, Jill Newby\textsuperscript{2},
Helen Christensen\textsuperscript{2}, Svetha Venkatesh\textsuperscript{1},
Alexis Whitton\textsuperscript{2}, Sunil Gupta\textsuperscript{1}\\
\textsuperscript{1}\textit{Applied Artificial Intelligence Initiative
(A\textsuperscript{2}I\textsuperscript{2}), Deakin University, Geelong,
VIC, Australia}\\
\textsuperscript{2}\textit{Black Dog Institute, University of New
South Wales, Sydney, NSW, Australia}}
\maketitle
\begin{abstract}
University students experience disproportionately high rates of common
mental health conditions, such as depression, which can impair learning,
social functioning, and overall well-being. Although lifestyle interventions
such as mindfulness and physical activity can reduce the symptoms,
many do not achieve symptomatic remission. Developing new approaches
to identify students with poor outcomes could enable earlier and more
targeted intervention.

Machine learning (ML) methods have increasingly been used to predict
remission in depressive patients. However, these ML models often suffer
from class imbalance, where there may be an unequal proportion of
people in the remitted group relative to the non-remitted group. This
imbalance can reduce model accuracy and bias predictions. To address
this, studies commonly employ the popular oversampling strategy SMOTE\foreignlanguage{american}{.
However, SMOTE has a notable limitation: it may generate invalid synthetic
minority samples. In a clinical context, these false positives can
lead to incorrect risk stratification, potentially delaying necessary
escalated care for patients unlikely to remit.}

\selectlanguage{american}%
In this paper, we introduce a novel and effective oversampling method
that addresses this shortcoming. Our approach leverages the variance
function of a Gaussian process to estimate the uncertainty of generated
minority samples to reduce false positives. We validate our method
on a depression dataset collected from university students and demonstrate
that it is better than existing oversampling approaches in predicting
remission (i.e., treatment outcome). By improving the reliable identification
of non-responders, our method provides a robust computational tool
to help clinicians rapidly pivot to adjunctive therapies, thereby
personalizing and optimizing mental health care pathways.
\end{abstract}

\section{Introduction\label{sec:Introduction}}

University students experience high rates of depression \foreignlanguage{american}{\cite{schwan2021perceptions,shvetcov2024passive}}.
Although evidence supports the effectiveness of various treatments,
the presence of prominent symptoms like \textit{amotivation} (loss
of motivational drive) and \textit{anhedonia} (reduced capacity to
experience pleasure) is often linked to poorer treatment responses
and worse long-term prognoses \cite{whitton2023distinct}. Lifestyle
interventions, such as physical activity and mindfulness, can serve
as primary or adjunctive care to reduce symptoms but even with the
delivery of treatment, non-remission rates remain problematically
high, ranging from 40\% to 52\% \cite{Blumenthal1999,eisendrath2016randomized,manger2019lifestyle}.
Because of this high rate of unsuccessful treatment, the early identification
of patients unlikely to achieve remission has become a critical priority
in both clinical practice and psychiatric research \cite{zhou2021machine,benoit2022using,wang2024examining,carr2025optimizing,calderon2026baseline,park2026prediction}.

\selectlanguage{american}%
Machine learning (ML) methods have achieved significant successes
across many healthcare domains, including mental health \cite{habehh2021machine,alanazi2022using,shvetcov2024passive,newby2025brief}.
Prior works have utilized popular ML methods such as \textit{k-nearest
neighbors} (kNN), \textit{decision tree} (DT), \textit{support vector
machine} (SVM), and \textit{random forest} (RF) to predict remission
outcomes in depression treatment \cite{zhou2021machine,benoit2022using,carr2025optimizing,calderon2026baseline}.
These predictive models typically rely on depression datasets encompassing
demographics, survey responses, psychiatric history, and treatment
types. However, accurately predicting remission remains a challenging
task because these training datasets are inherently imbalanced. Specifically,
the proportion of patients experiencing non-remission relative to
those achieving remission tends to be unequal, with literature indicating
minority group rates of only 30\% to 40\% \cite{rush2006acute,zhou2021machine,benoit2022using,park2026prediction}.

To rebalance training sets, recent studies \cite{kautzky2021combining,curtiss2024optimizing,shamshuzzoha2025novel,calderon2026baseline}
have increasingly leveraged a popular oversampling technique called
SMOTE\foreignlanguage{english}{ (Synthetic Minority Over-sampling
Technique) \cite{chawla2002smote}. SMOTE addresses class imbalance
by generating synthetic minority samples through the linear combination
of two real minority samples. It has proven to be comparable or superior
to other traditional methods (e.g., AdaSyn \cite{he2008adasyn}) and
deep learning approaches (e.g., CTGAN \cite{Xu2019} and TVAE \cite{Xu2019,borisov2022deep}).
However, \textit{SMOTE possesses a significant methodological weakness}.
Because SMOTE randomly interpolates a real minority sample with one
of its nearest neighbors, the newly generated minority sample may
inadvertently fall within the feature space of the majority class
if the selected neighbor is distant \cite{fernandez2018smote}. This
issue arises because SMOTE assumes the minority class region is convex,
whereas in reality, it may be non-convex. Consequently, SMOTE frequently
suffers from a high \textit{false-positive} rate, generating invalid
or incorrect minority samples \cite{chawla2002smote}.}

\selectlanguage{english}%
To address this limitation, we propose a novel method based on Gaussian
process (GP) \cite{rasmussen2003gaussian,Nguyen2020} to reduce the
false-positive rate of SMOTE. Unlike standard SMOTE and its existing
variants \cite{chawla2002smote,han2005borderline,nguyen2011borderline,sauglam2022novel},
our approach assigns a \textit{variance score} for each synthetic
minority sample using the \textit{variance function} of a GP. This
score effectively estimates the confidence levels (or uncertainty)
of the newly generated samples. We then filter out synthetic samples
with high variance scores (i.e., those exceeding a pre-specified threshold).
By doing so, we selectively retain only the synthetic minority samples
that are close to the true minority distribution, thereby significantly
reducing the generation of false positives. We refer to our method
as \textbf{SMOTE-VAR}.

We evaluate SMOTE-VAR using a clinical depression dataset comprising
784 university students across Australia \cite{huckvale2023protocol,newby2025brief}.
The dataset includes Depression, Anxiety, and Stress Scales-21 (DASS-21)
data and assigned treatment types. Each participant was assigned to
one of four treatment arms: digital mindfulness, digital sleep hygiene,
digital physical activity, or digital mood monitoring control treatment.
Furthermore, inspired by recent studies demonstrating the utility
of Global Positioning System (GPS) data for predicting mental health
statuses (such as stress \cite{shvetcov2024passive}, schizophrenia
\cite{jongs2020framework}, and depression \cite{muller2021depression}),
we extract mobility patterns as additional predictive features. Given
the dataset's class imbalance--64\% ``in remission'' versus 36\%
``non-remission''--we employ SMOTE-VAR to rebalance the training
data prior to training the predictive models.

In summary, our primary contributions are two-fold:
\begin{enumerate}
\item \textbf{SMOTE-VAR -- an effective oversampling method:} We propose
a novel GP-based oversampling approach that mitigates the generation
of false positives by filtering out highly uncertain synthetic minority
samples.
\item \textbf{Real-world clinical application:} We apply SMOTE-VAR alongside
five standard ML classifiers (kNN, DT, SVM, RF, and XGBoost) to predict
treatment remission in a real-world cohort of Australian university
students. Notably, the SVM classifier trained with SMOTE-VAR achieves
a \textit{Balanced Accuracy} (bACC) of 0.73 ($\pm$0.02), yielding
a 22\% improvement over a baseline SVM classifier trained without
oversampling.
\end{enumerate}
\selectlanguage{american}%
The remainder of this paper is organized as follows. Section \ref{sec:Related-Works}
summarizes literature on ML-based remission prediction and current
oversampling techniques, including traditional and deep learning approaches.
Section \ref{sec:Framework} details our primary methodological contribution,
SMOTE-VAR. Section \ref{sec:Experiments} describes the clinical dataset
and provides a comprehensive analysis of the experimental results.
Finally, Sections \ref{sec:Conclusion} and \ref{sec:Limitations-and-Future}
conclude the study, explain its limitations, and outline future research
directions.\selectlanguage{english}%

\section{Related Works\label{sec:Related-Works}}

\subsection{Remission Prediction with ML}

The application of machine learning (ML) to predict treatment remission
in patients with depression has gained significant traction, generally
falling into two methodological categories: non-oversampling and oversampling.
Non-oversampling approaches train predictive models directly on imbalanced
datasets without applying any rebalancing techniques \cite{zhou2021machine,benoit2022using,wang2024examining,park2026prediction}.
While methodologically straightforward, this approach often yields
suboptimal predictive performance, typically achieving a bACC of approximately
0.66--0.68. Conversely, oversampling strategies attempt to mitigate
class imbalance prior to model training, frequently utilizing the
Synthetic Minority Oversampling Technique (SMOTE) to rebalance the
training set \cite{kautzky2021combining,curtiss2024optimizing,shamshuzzoha2025novel,calderon2026baseline}.
By augmenting the training data with synthetic minority samples, these
models demonstrate improved performance \cite{calderon2026baseline}.
Despite this improvement, standard SMOTE exhibits a critical vulnerability:
the frequent generation of invalid or incorrect synthetic minority
samples (false positives). To address this inherent limitation, our
work introduces a novel oversampling framework.

\subsection{Imbalanced Classification}

Imbalanced classification challenges arise when the frequency of one
class vastly outnumbers the others within a training set. Oversampling
has emerged as a robust programmatic solution to this issue. While
there have been several methods proposed, such as ROSE \cite{menardi2014training}
and AdaSyn \cite{he2008adasyn}, the majority of existing oversampling
techniques are fundamental extensions of SMOTE \cite{chawla2002smote},
which generates synthetic minority instances by linearly interpolating
between two existing real minority samples. To combat SMOTE\textquoteright s
susceptibility to noise and outlier generation, several variants have
been proposed over the years \cite{batista2003balancing,batista2004study,han2005borderline,nguyen2011borderline}.
Alternative approaches leverage deep generative modeling, such as
the Conditional Tabular Generative Adversarial Network (CTGAN) \cite{Xu2019}
and Tabular Variational Autoencoders (TVAE) \cite{Xu2019,borisov2022deep},
which utilize generator or encoder networks to learn the distribution
of real minority samples. More recently, the capabilities of Large
Language Models (LLMs) have been adapted to address tabular data oversampling
\cite{Yang2024,nguyen2025large}.

However, within the mental health domain, deep learning-based oversampling
methods often underperform due to the characteristically small sample
sizes of clinical datasets. Simultaneously, traditional SMOTE-based
methods consistently fail to adequately regulate the generation of
false positives. Consequently, this paper presents a novel SMOTE-based
oversampling technique explicitly engineered to overcome this widespread
weakness.

\section{Framework\label{sec:Framework}}

This section outlines the mathematical framework of our study. We
first formalize the problem of imbalanced classification and the objective
of oversampling. Subsequently, we detail our proposed methodology,
SMOTE-VAR, which is designed to overcome the limitations of standard
interpolation techniques.

\subsection{Oversampling for Imbalanced Classification}

Let ${\cal D}_{train}=\{x_{i},y_{i}\}_{i=1}^{N}$ represent an \textit{imbalanced}
tabular dataset. Each instance comprises a feature vector $x_{i}$
with $M$ predictor variables $\{X_{1},...,X_{M}\}$ and a corresponding
target label $y_{i}$. We formulate the remission prediction as a
binary classification task, where $y_{i}\in\{0,1\}$. We designate
the class $Y=0$ as the \textit{majority} (\textit{negative}) class
and $Y=1$ as the \textit{minority} (\textit{positive}) class. The
subsets of majority and minority samples are denoted as ${\cal D}_{major}$
and ${\cal D}_{minor}$, respectively such that ${\cal D}_{train}={\cal D}_{major}\cup{\cal D}_{minor}$
and $\mid{\cal D}_{minor}\mid\ll\mid{\cal D}_{major}\mid$.

\begin{figure}[h]
\begin{centering}
\includegraphics[scale=0.5]{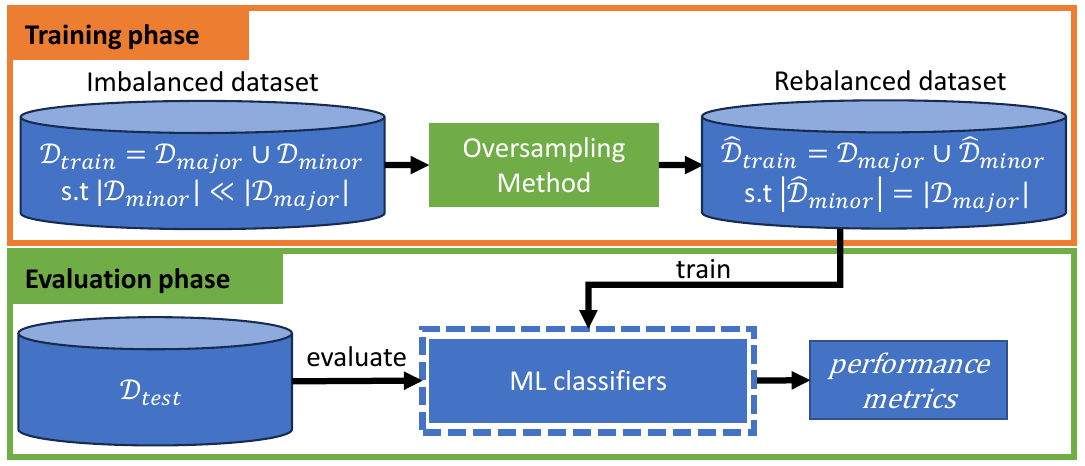}
\par\end{centering}
\caption{\label{fig:Training-and-evaluation}Training and evaluation phases
of an oversampling method. \textbf{Training:} the oversampling method
learns from the imbalanced dataset ${\cal D}_{train}$ to generate
synthetic minority samples $\hat{{\cal D}}_{minor}$ to construct
the rebalanced dataset $\hat{{\cal D}}_{train}$. \textbf{Evaluation:}
$\hat{{\cal D}}_{train}$ is used to train ML classifiers, and the
classifiers are evaluated on a held-out test set ${\cal D}_{test}$
to compute performance metrics (e.g., bACC). \textit{A higher score
implies a better oversampling method}.}
\end{figure}

The primary objective of an oversampling method is to learn a data
synthesizer from ${\cal D}_{train}$ capable of generating a set of
\textit{synthetic} minority samples, denoted as $\hat{{\cal D}}_{minor}$,
such that the classes are balanced ($\mid\hat{{\cal D}}_{minor}\mid=\mid{\cal D}_{major}\mid$).
These synthetic samples are then aggregated to construct a \textit{rebalanced}
training dataset, $\hat{{\cal D}}_{train}={\cal D}_{major}\cup\hat{{\cal D}}_{minor}$.
Ultimately, the efficacy of the oversampling method is evaluated by
training ML classifiers on $\hat{{\cal D}}_{train}$ and measuring
their \textit{Balanced Accuracy} (bACC) on a held-out test set ${\cal D}_{test}$.
A higher bACC indicates a more effective oversampling strategy. Figure
\ref{fig:Training-and-evaluation} illustrates the training and evaluation
phases for an oversampling method.

\subsection{The Proposed Method: SMOTE-VAR}

To address the limitations of existing techniques, we propose a novel
SMOTE-based oversampling framework, termed SMOTE-VAR.

\subsubsection{A probabilistic approach to reduce false positives}

Given a real minority sample $x_{i}$, the standard SMOTE algorithm
\cite{chawla2002smote} generates a synthetic minority sample $\hat{x}_{i}$
via linear interpolation:
\begin{equation}
\hat{x}_{i}=x_{i}+\lambda\times(x_{j}-x_{i}),\label{eq:SMOTE}
\end{equation}
where $x_{j}$ is a randomly selected neighbor from the \textit{k}-nearest
neighbors from the minority class and $\lambda\in(0,1)$ is a random
uniform variable.

\begin{figure}
\begin{centering}
\includegraphics[scale=0.6]{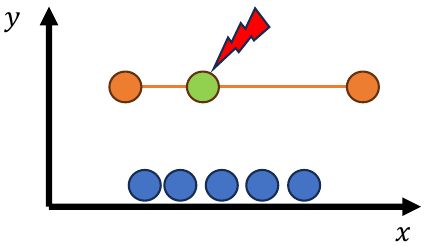}
\par\end{centering}
\caption{\label{fig:SMOTE-generates-false}SMOTE may generate an incorrect
synthetic minority sample (green dot) when the connecting line between
two real minority samples (orange dots) crosses the region of real
majority samples (blue dots).}
\end{figure}

While computationally efficient, SMOTE is highly susceptible to generating
false positive samples. As shown in Figure \ref{fig:SMOTE-generates-false},
SMOTE may generate an erroneous synthetic minority sample when the
line connecting two real minority samples inadvertently crosses the
region of majority samples. This geometric vulnerability occurs because
SMOTE implicitly assumes the minority class region is a convex set
\cite{fernandez2018smote}, whereas real-world clinical data distributions
are frequently non-convex.

Previous methodologies have attempted to mitigate this by identifying
and excluding ``distant'' neighbors from the interpolation process
\cite{han2005borderline,nguyen2011borderline}. However, this exclusionary
strategy inevitably generates an over-density of synthetic samples
immediately surrounding $x_{i}$ while completely neglecting the latent
feature space near the distant neighbor $x_{j}$. 

We propose a novel probabilistic strategy, termed \textbf{SMOTE-VAR},
which retains ``distant'' neighbors during interpolation but rigorously
evaluates the validity of the resulting synthetic samples. Standard
SMOTE uniformly assigns a hard minority label ($\hat{y}_{i}=1$) to
all generated samples $\hat{x}_{i}\in\hat{{\cal D}}_{minor}$ without
any measure of predictive confidence. We hypothesize that by calculating
a confidence measure for each assignment, we can systematically reject
synthetic samples exhibiting high uncertainty.

To estimate this uncertainty, we leverage the \textit{predictive variance}
of a Gaussian Process (GP) \cite{rasmussen2003gaussian,shahriari2016taking,Nguyen2020,Nguyen2021}
constructed over the spatial manifold of the true minority samples.
Because we are solely interested in measuring how well a newly generated
sample is supported by the surrounding true minority observations,
we do not require the GP to predict class labels. Instead, we compute
the posterior variance score $v_{\hat{x}}$ for every synthetic minority
sample $\hat{x}\in\hat{{\cal D}}_{minor}$:
\begin{align}
v_{\hat{x}} & =\sigma^{2}(\hat{x})=k(\hat{x},\hat{x})-\ensuremath{\mathbf{k}^{\text{T}}\mathbf{K}^{-1}\mathbf{k}},\label{eq:variance-score}
\end{align}
where $\sigma^{2}(\hat{x})$ denotes the GP variance function and
$k(\cdot,\cdot)$ is a kernel function. In this framework, we utilize
the \textit{Radial Basis Function} (RBF) kernel \cite{rasmussen2003gaussian,shahriari2016taking},
defined as $k(x_{i},x_{j})=\exp(-\|x_{i}-x_{j}\|^{2}/2\ell^{2})$,
where $\ell$ is the length-scale parameter. The RBF kernel is suited
for this task because it assumes spatial smoothness, causing the covariance
between points to decay exponentially with their distance. $\mathbf{k}$
is a vector with its \textit{i}-th element defined as $k(x_{i},\hat{x})$,
and $\mathbf{K}\in\mathbb{R}^{\mid{\cal D}_{minor}\mid\times\mid{\cal D}_{minor}\mid}$
represents the covariance matrix of the true minority data, where
its (\textit{i}, \textit{j})-th element is defined as $k(x_{i},x_{j})$.

Because the predictive variance $\sigma^{2}(\hat{x})$ is derived
from the covariance between data points, it serves as an effective
proxy for the uncertainty of the SMOTE assignment at $\hat{x}$. Uncertainty
naturally increases as $\hat{x}$ moves further away from the GP's
training data (the true minority samples). Consequently, we establish
a \textit{variance threshold} $\nu$. If $v_{\hat{x}}\leq\nu$, the
synthetic sample is deemed well-supported and retained. If $v_{\hat{x}}>\nu$,
the sample resides in a high-uncertainty, unsupported region of the
feature space and is discarded because it is likely to be invalid.
In practice, $\nu$ is set to a small value (e.g., $\nu\in[0.001,0.01]$)
to aggressively filter false positives while preserving a sufficient
volume of valid training samples. The sensitivity of the model to
$\nu$ is empirically evaluated in Section \ref{subsec:Impact-of-variance}.

\begin{figure}
\begin{centering}
\includegraphics[scale=0.6]{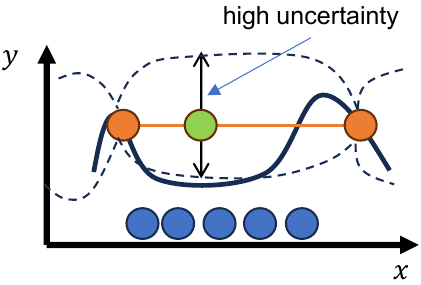}
\par\end{centering}
\caption{\label{fig:Our-VAR-component.}Our SMOTE-VAR. We generate a synthetic
minority sample (green dot) by linearly combining two real minority
samples (orange dots). This sample may be incorrect as it lies into
the region of real majority samples (blue dots). We compute a variance
score for this sample via a GP variance function. Because its variance
score is high (high uncertainty), the sample is rejected.}
\end{figure}

The GP is introduced solely as an uncertainty estimator during oversampling.
Once the filtered synthetic dataset is constructed, any downstream
classifier can be employed, preserving the classifier-agnostic nature
of SMOTE-VAR. Its conceptual mechanism is illustrated in Figure \ref{fig:Our-VAR-component.},
and it step-by-step implementation is presented in Algorithm \ref{alg:VAR-algorithm}.

\begin{algorithm}
\caption{\label{alg:VAR-algorithm}Our SMOTE-VAR algorithm.}

\LinesNumbered

\KwIn{real minority samples ${\cal D}_{minor}$}

\KwIn{synthetic minority samples $\hat{{\cal D}}_{minor}$}

\KwIn{variance threshold $\nu$}

\KwOut{filtered synthetic minority samples $\hat{{\cal D}}_{minor}^{*}$}

\Begin{

$\hat{{\cal D}}_{minor}^{*}=\emptyset$

fit a GP using real minority samples $x_{i}\in{\cal D}_{minor}$

\For{each $\hat{x}_{i}\in\hat{{\cal D}}_{minor}$ }{

compute its \textit{variance score} $v_{\hat{x}_{i}}$ with Eq. (\ref{eq:variance-score})

\If{$v_{\hat{x}_{i}}\leq\nu$}{

$\hat{{\cal D}}_{minor}^{*}=\hat{{\cal D}}_{minor}^{*}\cup\{\hat{x}_{i}\}$

}

}

}
\end{algorithm}

\subsubsection{Discussion}

Unlike prior distance-based filtering techniques \cite{han2005borderline,nguyen2011borderline},
SMOTE-VAR does not aggressively exclude distant neighbors, thereby
preventing the generation of over-dense, localized clusters of synthetic
data. By probabilistically evaluating the correctness of a synthetic
sample based on GP uncertainty, our method successfully spans the
interpolation space, generating valid samples near both $x_{i}$ and
$x_{j}$. SMOTE-VAR only selectively rejects synthetic instances that
land in high-variance, sparsely populated regions far from any true
minority data, as shown in Figure \ref{fig:Our-VAR-component.}.

A critical advantage of utilizing GP variance over simple heuristic
distance metrics (e.g., Euclidean distance) is its capacity to act
as a globally aware, non-parametric uncertainty estimator. Traditional
distance-based filtering evaluates synthetic samples in isolation,
relying solely on pairwise proximity while fundamentally ignoring
the structural topology of the feature space. In contrast, the GP
variance score dynamically incorporates local sample density, kernel
smoothness, and the complex spatial correlations among all true minority
samples. Because the inverse covariance matrix $\mathbf{K}^{-1}$
captures the collective spatial distribution of the minority class,
the resulting variance calculation intrinsically maps the underlying
non-linear manifold. Consequently, a synthetic sample generated in
a sparse, unsupported region will correctly register a critically
high variance, whereas a sample generated at the exact same Euclidean
distance but within a densely populated, highly correlated region
will be validated. This structural awareness ensures that the filtering
mechanism adapts to the local geometry of the clinical data, making
GP variance a more robust validation measure than rigid, localized
distance thresholds.

\section{Experiments\label{sec:Experiments}}

\selectlanguage{american}%
This section details the empirical evaluation of SMOTE-VAR. We outline
the clinical dataset, the preprocessing pipeline, and the experimental
configurations, followed by a comprehensive analysis of the predictive
performance and ablation studies.

\subsection{Dataset and Feature Engineering}

\subsubsection{Clinical cohort and pre-processing}

The dataset was derived from the Vibe-Up study \cite{huckvale2023protocol,newby2025brief},
a clinical trial encompassing university students exhibiting elevated
symptoms of psychological distress. Data were collected in the context
of an adaptive clinical trial involving 12 sequential mini trials,
all run between 2021-2023. From an initial cohort of 1,282 participants,
784 individuals provided concurrent GPS mobility data. Based on \cite{lovibond1995manual},
scores that fall below the thresholds of 9 for Depression, 7 for Anxiety,
and 15 for Stress indicate that an individual is ``in remission''.
For illustrative purposes, we present some \textit{fake examples}
of the depression dataset and corresponding GPS logs in Table \ref{tab:Example-GPS-data}.

\begin{table}[h]
\caption{\label{tab:Example-GPS-data}Some \textbf{illustrative examples} of
our dataset. Table (a) shows two students along with their DASS scores,
treatment types, and treatment outcomes (i.e., labels). Each student
has a unique ID. Table (b) shows their visited locations (in terms
of latitude and longitude). Each location is associated with a timestamp.}

\begin{centering}
\subfloat[Students with DASS scores, treatment types, and labels. \textquotedblleft DASS\_bl\textquotedblright{}
and \textquotedblleft DASS\_pre\textquotedblright{} stand for DASS
baseline and DASS pre-treatment.]{
\centering{}%
\begin{tabular}{|l|l|l|l|l|l|l|l|r|}
\hline 
\textbf{Student} & \foreignlanguage{english}{\textbf{DASS\_bl1}} & \foreignlanguage{english}{\textbf{...}} & \foreignlanguage{english}{\textbf{DASS\_bl21}} & \foreignlanguage{english}{\textbf{DASS\_pre1}} & \foreignlanguage{english}{\textbf{...}} & \foreignlanguage{english}{\textbf{DASS\_pre21}} & \foreignlanguage{english}{\textbf{Treatment}} & \foreignlanguage{english}{\textbf{Outcome}}\tabularnewline
\hline 
\hline 
abcxyz12 & \foreignlanguage{english}{1} & \foreignlanguage{english}{...} & \foreignlanguage{english}{0} & \foreignlanguage{english}{2} & \foreignlanguage{english}{...} & \foreignlanguage{english}{3} & mindfulness & \foreignlanguage{english}{remission}\tabularnewline
\hline 
\foreignlanguage{english}{12abc456} & \foreignlanguage{english}{2} & \foreignlanguage{english}{...} & \foreignlanguage{english}{1} & \foreignlanguage{english}{3} & \foreignlanguage{english}{...} & \foreignlanguage{english}{1} & physical activity & \foreignlanguage{english}{non-remission}\tabularnewline
\hline 
\end{tabular}}
\par\end{centering}
\centering{}\subfloat[Students with GPS locations.]{
\centering{}%
\begin{tabular}{|l|l|r|r|}
\hline 
\textbf{Student} & \textbf{Timestamp} & \textbf{Latitude} & \textbf{Longitude}\tabularnewline
\hline 
\hline 
abcxyz12 & 15/11/2021 07:35:00 & -30.8036 & 118.5869\tabularnewline
\hline 
abcxyz12 & 15/11/2021 07:55:05 & -30.8879 & 118.5271\tabularnewline
\hline 
abcxyz12 & 16/11/2021 08:36:39 & -30.9712 & 118.6020\tabularnewline
\hline 
\foreignlanguage{english}{12abc456} & 19/11/2021 16:11:31 & -17.4219 & 142.9478\tabularnewline
\hline 
\end{tabular}}
\end{table}

\begin{figure}[h]
\selectlanguage{english}%
\begin{centering}
\includegraphics[scale=0.5]{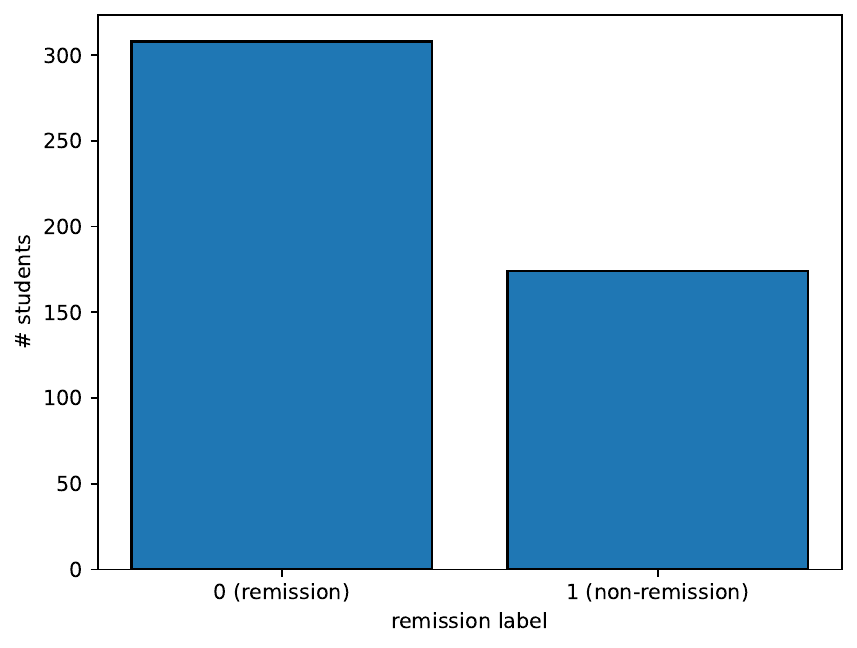}
\par\end{centering}
\caption{\label{fig:Distributions-of-amotivation}Class distribution of the
refined dataset. Among 482 students, 64\% of them are \textquotedblleft in
remission\textquotedblright{} and 36\% are \textquotedblleft non-remission\textquotedblright .}
\selectlanguage{english}%
\end{figure}

To ensure spatial data integrity, we applied the following strict
filtering protocols \cite{palmius2016detecting,raugh2020geolocation}:
excluding inaccurate GPS coordinates with an accuracy radius exceeding
35 meters, and removing duplicate spatial logs defined by a Haversine
distance of less than 500 meters. Participants with fewer than two
days of recorded mobility data were also excluded, resulting in a
refined analytical cohort of 482 students. The final dataset exhibits
a pronounced class imbalance, with 308 students (64\%) achieving treatment
remission and 174 students (36\%) classified as non-remission. Figure
\ref{fig:Distributions-of-amotivation} displays the distributions
of remission labels in our refined dataset.

\subsubsection{Feature extraction and imputation}

\selectlanguage{english}%
The predictive feature space was derived from clinical survey responses,
intervention assignments, and continuous GPS mobility logs. To establish
robust, aggregate measures of psychological distress, we computed
the sum of the 21 individual Depression, Anxiety, and Stress Scales
(DASS) items to derive a total baseline score (denoted as \textit{DASS\_bl\_total})
and a total pre-treatment score (denoted as \textit{DASS\_pre\_total}).
Furthermore, to capture digital behavioral phenotypes, we extracted
two daily spatial features from the GPS logs \foreignlanguage{american}{\cite{barnett2018relapse,raugh2020geolocation,muller2021depression}}:
\textit{the number of daily locations visited} ($n_{location}$) and
\textit{the average daily distance traveled} (\textbf{$dist_{travel}$}).
For example, as detailed in Table \ref{tab:Example-GPS-data}(b),
student ``abcxyz12'' registered $n_{location}=2$ and $dist_{travel}=4.01$
on 15/11/2021.

\begin{figure}[th]
\selectlanguage{english}%
\begin{centering}
\subfloat[]{\begin{centering}
\includegraphics[scale=0.28]{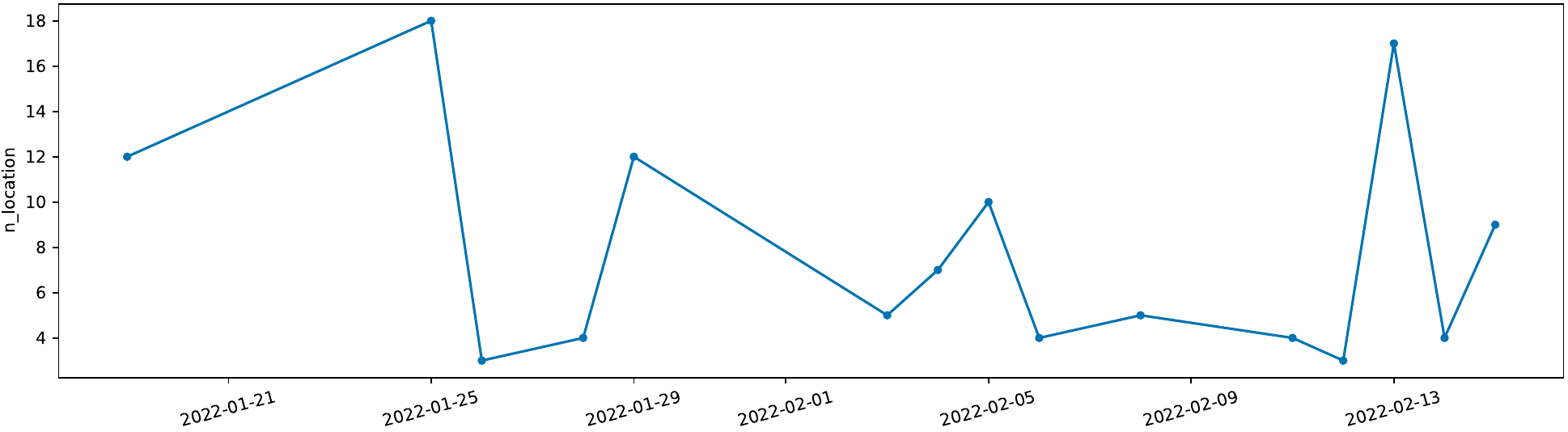}
\par\end{centering}
}\subfloat[]{\begin{centering}
\includegraphics[scale=0.28]{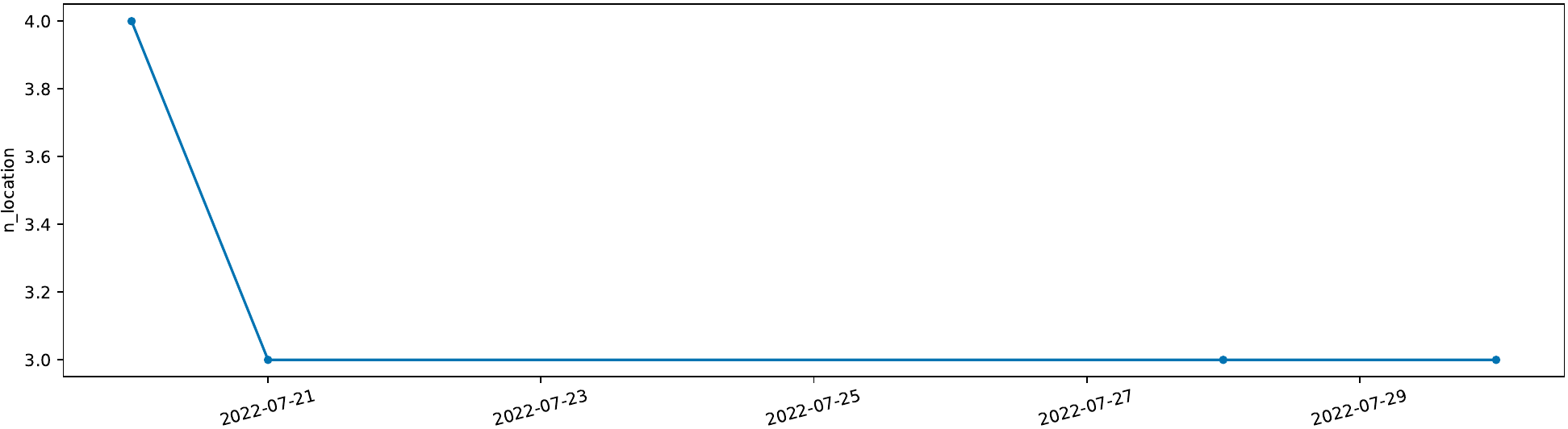}
\par\end{centering}
}
\par\end{centering}
\caption{\label{fig:Mobility-patterns}For an illustrative purpose, we show
\textbf{illustrative mobility patterns} of two students. The x-axis
shows the date while the y-axis shows the number of visited locations
in one day.}
\selectlanguage{english}%
\end{figure}

\selectlanguage{american}%
Because the temporal length of mobility data varied significantly
across participants--as visualized in the divergent mobility patterns
of two students in Figure \ref{fig:Mobility-patterns}--we applied
linear imputation \cite{kazijevs2023deep} to standardize the spatial
feature trajectories to a uniform 15-day observation window, ensuring
consistent mathematical dimensionality across all samples for subsequent
model training.

\subsection{Experimental Settings}

\subsubsection{Baselines and model configurations}

We benchmarked SMOTE-VAR against a robust suite of 10 methodologies:
standard non-oversampling (\textit{Imbalance}), traditional interpolation
methods (SMOTE \cite{chawla2002smote}, SMOTE-NC \cite{fernandez2018smote},
AdaSyn \cite{he2008adasyn}, and several SMOTE variants \cite{batista2003balancing,batista2004study,han2005borderline,nguyen2011borderline}),
deep generative models (CTGAN \cite{Xu2019} and TVAE \cite{Xu2019,borisov2022deep}),
and Large Language Model-based generation (ImbLLM \cite{nguyen2025large}).
For our proposed SMOTE-VAR framework, the variance threshold was established
at $\nu=0.001$.

\subsubsection{Classifiers and validation strategy}

Following previous works \cite{zhou2021machine,benoit2022using,wang2024examining,carr2025optimizing,calderon2026baseline},
the rebalanced datasets were utilized to train five standard machine
learning classifiers: \textit{k-Nearest Neighbors} (kNN), \textit{Support
Vector Machine} (SVM), \textit{Decision Tree} (DT), \textit{Random
Forest} (RF), and \textit{XGBoost} (XGB). We tuned their hyper-parameters
(detailed in Table \ref{tab:Hyper-parameters-of-ML}) using five-fold
cross-validation on the training sets. To rigorously evaluate generalization,
the dataset was subjected to a randomized 90/10 train-test split,
repeated across ten independent random seeds. Model performance is
reported as the average \textit{Balanced Accuracy} (bACC) alongside
its standard deviation. Balanced Accuracy (bACC) is defined as: $\text{bACC}=\frac{\text{Sensitivity}+\text{Specificity}}{2}$,
where $\text{bACC}\in[0,1]$ and \textit{higher score is better}.
We also report other performance metrics such as F1-score, Area Under
the Curve (AUC) score, Sensitivity, Specificity, and Receiver Operating
Characteristic (ROC) curve.

\begin{table}[h]
\selectlanguage{english}%
\caption{\label{tab:Hyper-parameters-of-ML}Hyper-parameters of ML classifiers
to predict treatment remission.}

\centering{}%
\begin{tabular}{|c|V{\linewidth}|}
\hline 
\textbf{Classifier} & \textbf{Hyper-parameters}\tabularnewline
\hline 
\hline 
kNN & \textit{n\_neighbors}: \{1, 2, ..., 10\}\tabularnewline
\hline 
SVM & \textit{kernel}: \{linear, rbf\}

\textit{gamma}: \{0.1, 0.5, 1.0\}

\textit{C}: \{0.1, 0.5, 1.0, 10.0\}\tabularnewline
\hline 
DT & \textit{max\_depth}: \{1, 2, ..., 12\}\tabularnewline
\hline 
RF & \textit{n\_estimators}: \{1, 10, 50, 100\}

\textit{max\_depth}: \{1, 2, ..., 12\}

\textit{max\_features}: \{sqrt, log2, None\}\tabularnewline
\hline 
XGB & \textit{n\_estimators}: \{1, 10, 50, 100\}

\textit{max\_depth}: \{1, 2, ..., 12\}

\textit{max\_features}: \{sqrt, log2, None\}\tabularnewline
\hline 
\end{tabular}\selectlanguage{english}%
\end{table}

\subsection{Results and Discussions}

Table \ref{tab:Main-experiments} reports bACC of each oversampling
method combined with five ML classifiers.

\selectlanguage{english}%
\begin{table}[h]
\caption{\label{tab:Main-experiments}bACC $\pm$ (standard deviation) of each
oversampling method combined with five popular ML classifiers. \textbf{Bold}
and \uline{underline} indicate the best and second-best methods.}

\centering{}%
\begin{tabular}{|l|cccccccc|}
\hline 
\rowcolor{header_color}bACC & Imbalance & AdaSyn & SMOTE & SMOTE-NC & CTGAN & TVAE & ImbLLM & SMOTE-VAR\tabularnewline
\hline 
\hline 
\multirow{2}{*}{kNN} & 0.66 & 0.65 & \textbf{0.68} & \textbf{0.68} & 0.66 & \uline{0.67} & 0.66 & \uline{0.67}\tabularnewline
 & (0.02) & (0.02) & (0.02) & (0.01) & (0.02) & (0.02) & (0.02) & (0.01)\tabularnewline
\hline 
\rowcolor{even_color}SVM & 0.51 & \textbf{0.73} & \uline{0.72} & \textbf{0.73} & 0.66 & \textbf{0.73} & \textbf{0.73} & \textbf{0.73}\tabularnewline
\rowcolor{even_color} & (0.00) & (0.02) & (0.02) & (0.02) & (0.02) & (0.02) & (0.02) & (0.02)\tabularnewline
\hline 
DT & 0.62 & 0.59 & 0.62 & 0.61 & 0.59 & 0.62 & \uline{0.64} & \textbf{0.65}\tabularnewline
 & (0.02) & (0.03) & (0.02) & (0.03) & (0.01) & (0.02) & (0.02) & (0.02)\tabularnewline
\hline 
\rowcolor{even_color}RF & 0.61 & 0.60 & 0.61 & 0.60 & 0.60 & \uline{0.63} & \uline{0.63} & \textbf{0.65}\tabularnewline
\rowcolor{even_color} & (0.02) & (0.03) & (0.02) & (0.03) & (0.01) & (0.02) & (0.02) & (0.02)\tabularnewline
\hline 
XGB & 0.59 & 0.58 & 0.62 & 0.59 & 0.59 & \uline{0.63} & 0.60 & \textbf{0.65}\tabularnewline
 & (0.01) & (0.03) & (0.01) & (0.02) & (0.01) & (0.02) & (0.02) & (0.02)\tabularnewline
\hline 
\rowcolor{childheader_color}Average & \textcolor{red}{0.60} & \textcolor{red}{0.63} & \textcolor{red}{0.65} & \textcolor{red}{0.64} & \textcolor{red}{0.62} & \textcolor{red}{\uline{0.66}} & \textcolor{red}{0.65} & \textbf{\textcolor{red}{0.67}}\tabularnewline
\hline 
\end{tabular}
\end{table}

\selectlanguage{american}%
Our method SMOTE-VAR consistently achieved the highest average balanced
accuracy. Across five ML classifiers, it was best performing with
four classifiers and second-best with another classifier. Its average
improvement over TVAE (the runner-up method) was 1\% and SMOTE (the
most popular baseline) was 2\%. More importantly, it yielded a substantial
7\% improvement over Imbalance (the method without oversampling).
Although SMOTE-VAR achieved the highest average balanced accuracy
across the evaluated classifiers, paired Wilcoxon statistical tests
against the strongest baselines (TVAE and SMOTE) did not reveal statistically
significant differences ($p>0.05$). This is likely attributable to
the relatively small performance gap (approximately 1-2\%) and the
limited number of repeated random splits.

\selectlanguage{english}%
All oversampling methods were much better than Imbalance. Interestingly,
traditional interpolation methods (e.g., AdaSyn, SMOTE, and SMOTE-NC)
demonstrated superior efficacy compared to the highly complex CTGAN
model. TVAE and ImbLLM served as the closest competitors, while SMOTE-VAR
achieved the highest average balanced accuracy. Notably, the SVM classifier
paired with SMOTE-VAR achieved a bACC of 0.73, representing a 22\%
improvement over standard classifiers trained on imbalanced data.
In summary, our method SMOTE-VAR achieved the highest average performance,
maintaining consistency across multiple ML classifiers. \textit{Because
SVM was the most effective classifier}, \textit{it is the default
classifier for our following experiments}.

\begin{figure}[th]
\begin{centering}
\includegraphics[scale=0.5]{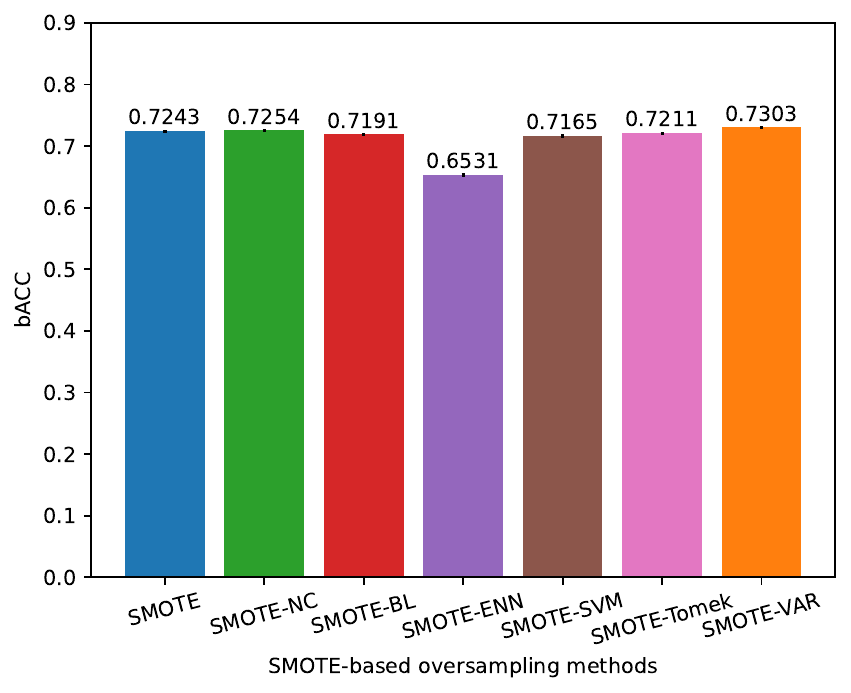}
\par\end{centering}
\caption{\label{fig:AUC-scores-of-SMOTE-variants}bACC of our method SMOTE-VAR
and other SMOTE variants.}

\end{figure}

\textbf{Comparison with SMOTE variants.} Because our method is based
on SMOTE, we also compared it with SMOTE variants, including SMOTE-BL
\cite{han2005borderline}, SMOTE-ENN \cite{batista2004study}, SMOTE-SVM
\cite{nguyen2011borderline}, and SMOTE-Tomek \cite{batista2003balancing}.
Figure \ref{fig:AUC-scores-of-SMOTE-variants} shows that SMOTE-VAR
maintained the highest overall predictive performance while SMOTE-NC
was the second-best method. Other SMOTE-based methods, except SMOTE-ENN,
behaved similarly.

\selectlanguage{american}%

\subsection{Ablation studies}

To unpack the mechanics of SMOTE-VAR, we conducted targeted ablation
studies analyzing temporal robustness, variance threshold sensitivity,
and feature importance.

\subsubsection{Temporal robustness}

\begin{figure}[h]
\selectlanguage{english}%
\begin{centering}
\includegraphics[scale=0.5]{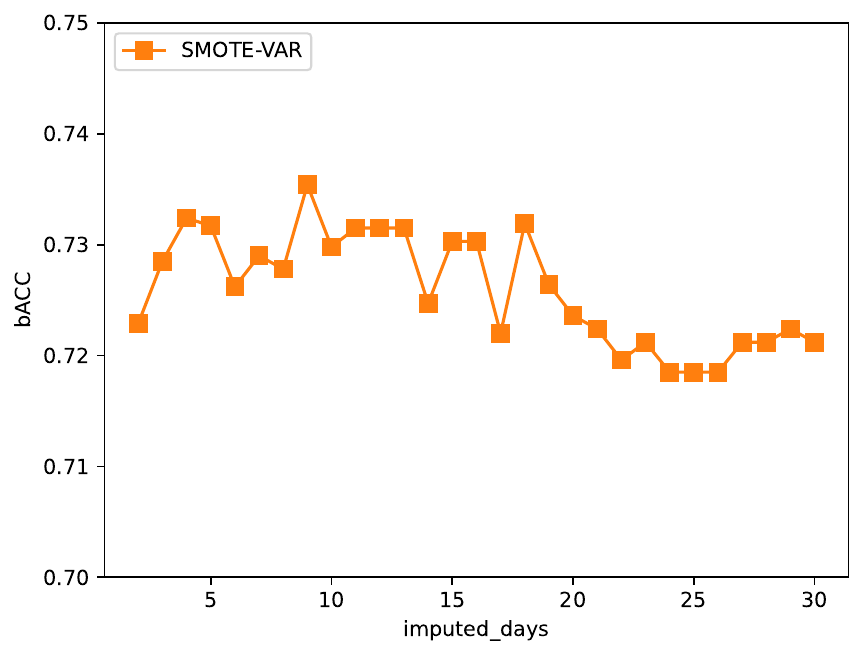}
\par\end{centering}
\caption{\label{fig:AUC-score-vs.-imputed-days}bACC vs. the number of imputed
days.}
\selectlanguage{english}%
\end{figure}

As presented in Figure \ref{fig:AUC-score-vs.-imputed-days}, variations
in the imputed observation window (ranging from 2 to 30 days) yielded
negligible fluctuations in predictive accuracy, with bACC scores remaining
robustly above 0.72.

\subsubsection{Imputation method}

Table \ref{tab:AUC-score-vs.-imputation} shows our performance with
different imputation methods, highlighting that ``linear'' and ``s-linear''
functions achieved the highest efficacy. Other imputation methods
also performed well.

\begin{table}[h]
\selectlanguage{english}%
\caption{\label{tab:AUC-score-vs.-imputation}bACC vs. imputation methods.}

\begin{centering}
\begin{tabular}{|l|c|}
\hline 
\textbf{Imputation method} & \textbf{bACC}\tabularnewline
\hline 
\hline 
linear & \textbf{0.7303 }(0.02)\tabularnewline
\hline 
nearest & 0.7231 (0.02)\tabularnewline
\hline 
nearest-up & 0.7204 (0.02)\tabularnewline
\hline 
zero & \uline{0.7266} (0.02)\tabularnewline
\hline 
s-linear & \textbf{0.7303 }(0.02)\tabularnewline
\hline 
previous & \uline{0.7266} (0.02)\tabularnewline
\hline 
next & 0.7215 (0.02)\tabularnewline
\hline 
\end{tabular}
\par\end{centering}
\selectlanguage{english}%
\end{table}

\subsubsection{Impact of variance threshold\label{subsec:Impact-of-variance}}

\begin{figure}[th]
\selectlanguage{english}%
\begin{centering}
\includegraphics[scale=0.5]{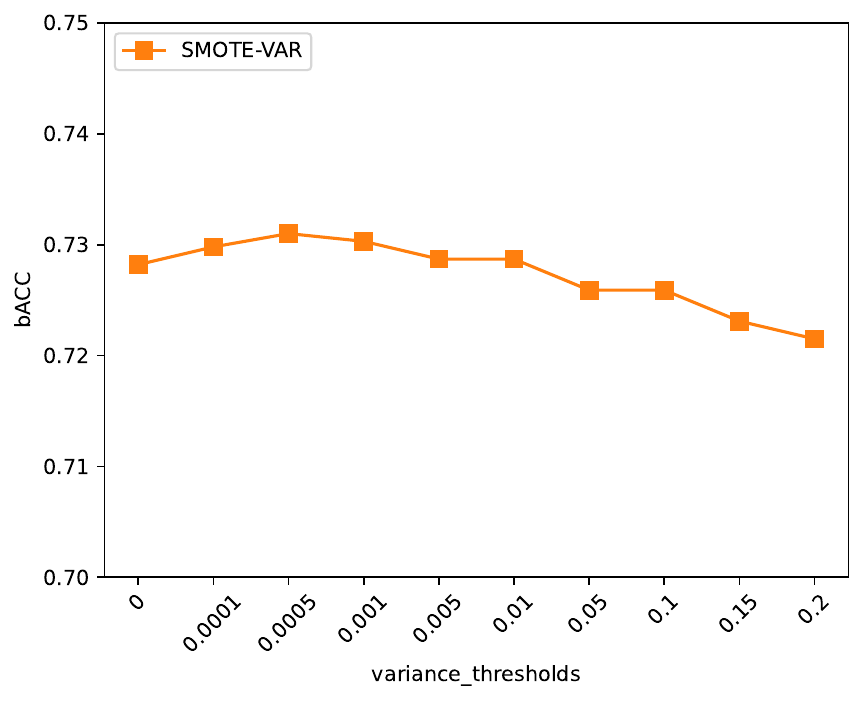}
\par\end{centering}
\caption{\label{fig:AUC-score-vs.-variance}bACC vs. variance threshold $\nu$.}
\selectlanguage{english}%
\end{figure}

Modulating the variance threshold $\nu$ highlighted the critical
balance between sample retention and false-positive rejection. We
present our bACC scores across different thresholds $\nu$ in Figure
\ref{fig:AUC-score-vs.-variance}. Optimal performance was observed
within the tight threshold bounds of $\nu\in[0.001,0.01]$. Thresholds
exceeding 0.05 diminished bACC scores by permitting the inclusion
of false positives, while overly restrictive thresholds ($\nu<0.0005$)
indiscriminately discarded valid synthetic samples, resulting in insufficient
training volume.

\subsubsection{Feature importance}

\begin{figure}[th]
\selectlanguage{english}%
\begin{centering}
\includegraphics[scale=0.5]{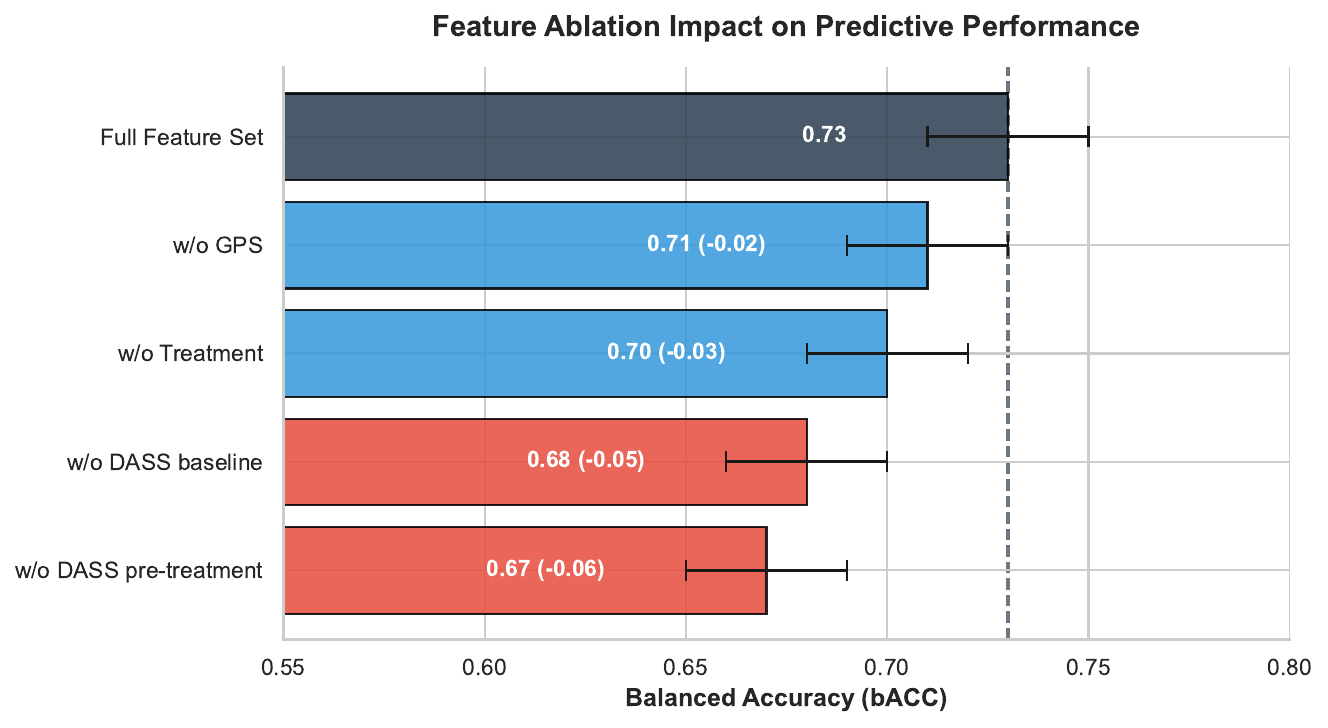}
\par\end{centering}
\caption{\label{fig:AUC-score-vs.-feature.}bACC vs. features.}

\selectlanguage{english}%
\end{figure}

To determine predictive drivers, we documented the bACC drops when
systematically removing features in Figure \ref{fig:AUC-score-vs.-feature.}.
The results indicated that subjective clinical assessments--specifically
DASS pre-treatment and baseline scores--are the primary drivers of
model accuracy. The systematic removal of these feature domains precipitated
bACC reductions of approximately 5\% to 6\%. Conversely, the exclusion
of GPS mobility features resulted in a marginal 2\% decline in the
overall bACC score. This suggests that while digital phenotyping via
spatial behavior provides supplementary predictive value, direct clinical
symptomatology remains paramount for forecasting depression remission.

Following \cite{misiuk2024multivariate}, we further present the calculated
\textit{importance score} of each individual feature. We expect that
removing an important feature should have a significant change in
the classifier predictions whereas the absence of an unimportant feature
should have little effect. Given a test set ${\cal D}_{test}=\{x_{j}\}_{j=1}^{N'}$,
let $f(x_{j})$ and $f_{i}(x_{j})$ be the predictions of the original
classifier (i.e., using all features) and the modified classifier
(i.e., removing one feature $X_{i}$).

The \textit{importance score} of a feature $X_{i}$ is computed as:
\begin{equation}
s_{X_{i}}=\frac{\frac{1}{N'}\sum_{j=1}^{N'}\mid f(x_{j})-f_{i}(x_{j})\mid}{\frac{1}{N'}\sum_{j=1}^{N'}\mid f(x_{j})-\mu_{f}\mid},\label{eq:importance-score}
\end{equation}
where $N'$ is the number of samples in the test set ${\cal D}_{test}$
and $\mu_{f}=\frac{1}{N'}\sum_{j=1}^{N'}f(x_{j})$ is the mean value
of the predictions of the original classifier. Since the importance
score indicates the deviation from the original predictions, a higher
value for $s_{X_{i}}$ means the feature $X_{i}$ is more important.

\begin{figure}[h]
\selectlanguage{english}%
\begin{centering}
\includegraphics[scale=0.5]{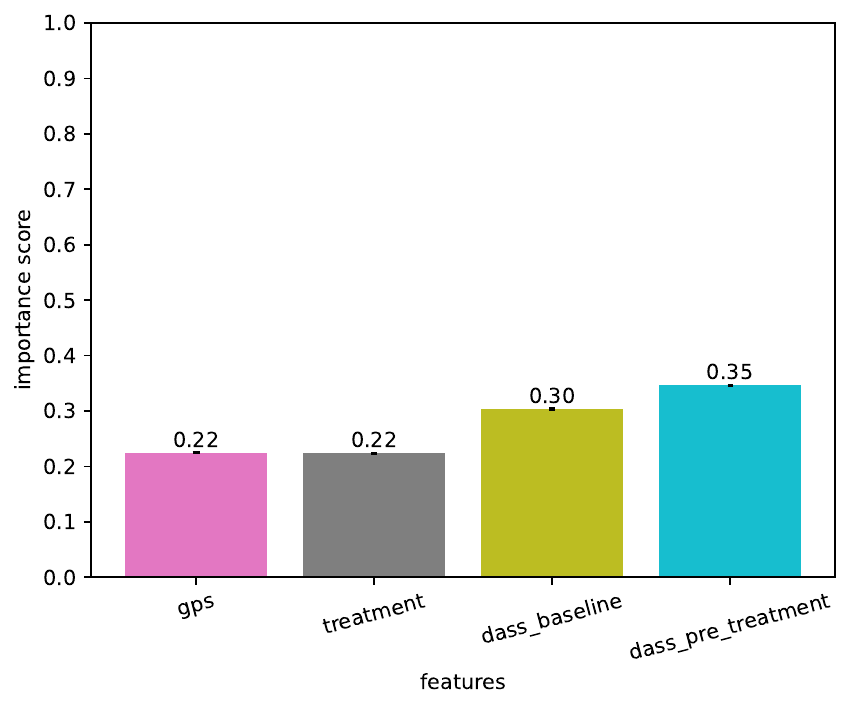}
\par\end{centering}
\caption{\label{fig:Importance-score}Feature importance. \textit{A higher
score indicates a more important feature}.}
\selectlanguage{english}%
\end{figure}

From Figure \ref{fig:Importance-score}, DASS pre-treatment has the
highest score, indicating it is the most important feature, following
by DASS baseline. These results also agree with the results in Figure
\ref{fig:AUC-score-vs.-feature.}.

\subsubsection{Other performance metrics}

\begin{figure}[H]
\selectlanguage{english}%
\begin{centering}
\includegraphics[scale=0.5]{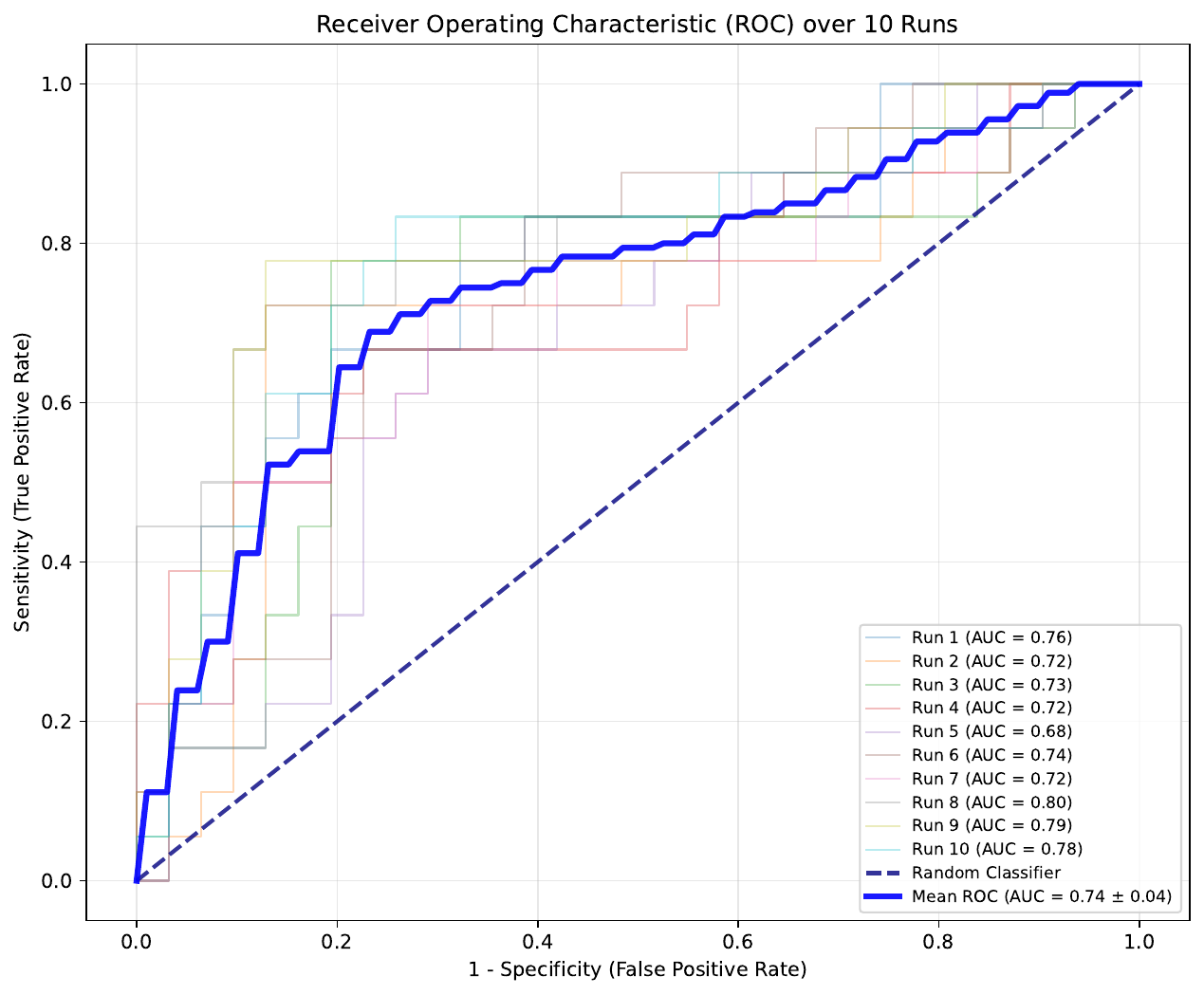}
\par\end{centering}
\caption{\label{fig:Receiver-Operating-Characteristi}Receiver Operating Characteristic
(ROC) of SMOTE-VAR over 10 runs.}
\selectlanguage{english}%
\end{figure}

To provide a comprehensive evaluation of the predictive stability
and clinical utility of SMOTE-VAR, we further present its performance
across multiple metrics in Figures \ref{fig:AUC-score,-F1-score,-Sensitivity}
and \ref{fig:Receiver-Operating-Characteristi}.

As detailed in Figure \ref{fig:AUC-score,-F1-score,-Sensitivity},
the SVM classifier paired with SMOTE-VAR achieved a mean AUC of 0.74
and a mean F1-score of 0.66 across the 10 independent test runs. Most
notably, the model demonstrated a mean Specificity of 0.82 and a mean
Sensitivity of 0.64. In the context of this study, where non-remission
is designated as the positive minority class, this high specificity
indicates that the model is highly reliable at correctly identifying
patients who \textit{will} successfully achieve remission (true negatives).
Simultaneously, the sensitivity of 64\% demonstrates a robust ability
to flag the harder-to-predict non-remitters. From a clinical informatics
perspective, this balance is highly practical: it minimizes false
alarms for patients on track to recover, while successfully catching
the majority of at-risk students who may require rapid escalation
to adjunctive therapies.

Figure \ref{fig:Receiver-Operating-Characteristi} visualizes the
Receiver Operating Characteristic (ROC) curves for each individual
run, alongside the aggregated mean ROC curve. The tight clustering
of the individual curves around the mean--reflected by the narrow
standard deviation of the AUC (0.74$\pm$0.04)--highlights the stability
and generalization capability of the SMOTE-VAR framework. This consistency
across random data splits further validates our hypothesis: by probabilistically
filtering out high-variance synthetic samples, SMOTE-VAR effectively
stabilizes the decision boundary and prevents the classifier from
overfitting to the noisy interpolation artifacts common in standard
oversampling techniques.

\begin{figure}[th]
\selectlanguage{english}%
\begin{centering}
\includegraphics[scale=0.5]{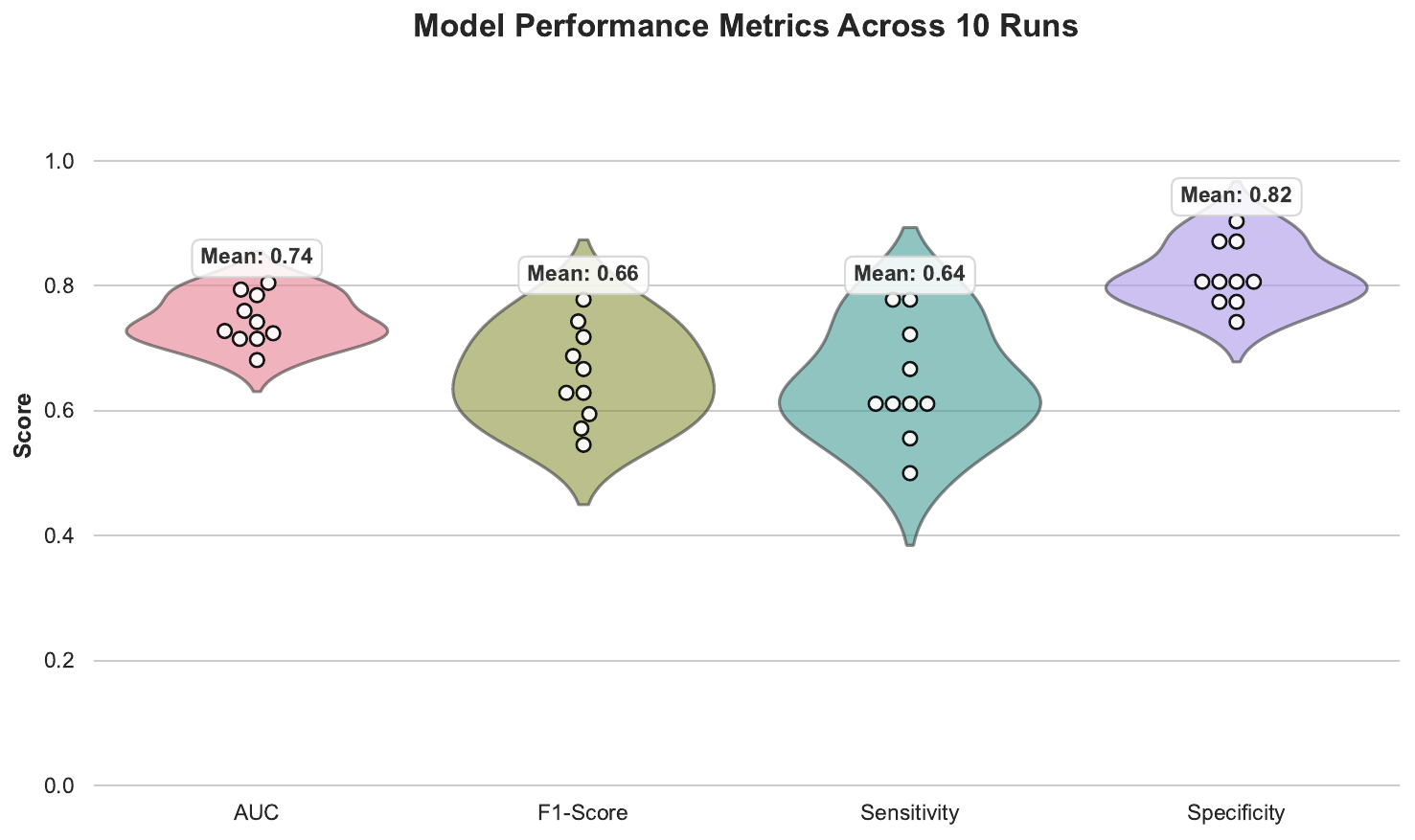}
\par\end{centering}
\caption{\label{fig:AUC-score,-F1-score,-Sensitivity}AUC, F1-score, Sensitivity,
and Specificity of SMOTE-VAR across 10 runs.}

\selectlanguage{english}%
\end{figure}

\selectlanguage{english}%

\section{Conclusion\label{sec:Conclusion}}

In this study, we demonstrate that the integration of clinical survey
data and GPS location patterns can effectively predict treatment remission
in depressive students. This predictive capability offers valuable
insights for the development of targeted, early-stage mental health
interventions. Departing from previous approaches, we introduce a
novel and effective oversampling technique to rebalance imbalanced
training datasets in machine learning-based remission prediction.
Our method incorporates a crucial probabilistic component into standard
SMOTE, assigning an uncertainty score to each synthetic minority sample
to systematically reduce false positives. We verify the effectiveness
of our approach on a real-world depression dataset collected from
Australian students, where it consistently achieved the highest average
predictive performance among the evaluated oversampling methods.

\section{Limitations and Future Directions\label{sec:Limitations-and-Future}}

While this study demonstrates the efficacy of SMOTE-VAR in predicting
treatment remission, several methodological and clinical limitations
should be noted to guide future research.

First, from a \textbf{methodological perspective}, scaling the SMOTE-VAR
framework to highly complex datasets presents computational challenges.
Standard Gaussian Process covariance functions are susceptible to
the curse of dimensionality, losing discriminatory power when applied
to highly dimensional data. Furthermore, exact GP inference scales
at $O(N^{3})$, which can introduce computational bottlenecks when
the minority class is exceptionally large. Future iterations of this
framework should integrate \textit{Deep Kernel Learning} (DKL) \cite{wilson2016deep}
to project high-dimensional inputs into a lower-dimensional latent
space, and substitute exact GPs with \textit{Sparse Gaussian Processes}
\cite{snelson2005sparse} utilizing pseudo-inputs to reduce computational
complexity to $O(NM^{2})$. Additionally, incorporating \textit{Automatic
Relevance Determination} (ARD) \cite{wipf2007new} will better equip
the model to handle the heterogeneous mix of continuous and categorical
variables native to clinical data.

Second, regarding the \textbf{clinical study design}, the primary
outcome of remission was assessed straight after a brief, 2-week digital
mental health intervention. In standard psychiatric literature, the
full clinical benefits of lifestyle interventions, such as physical
activity and mindfulness, typically manifest over a longer duration,
often between 12 to 16 weeks. Evaluating outcomes after a 2-week intervention
may primarily capture early remitters rather than long-term remission,
potentially contributing to the high prevalence of the non-remission
class in our dataset. Future studies should evaluate the predictive
performance of SMOTE-VAR over extended longitudinal follow-ups to
capture a more complete picture of treatment efficacy.

Finally, regarding \textbf{data processing}, our feature importance
analysis revealed that GPS mobility data contributed marginally to
the model's predictive power, resulting in only a 2\% decrease in
the balanced accuracy (bACC) score when removed. This may be partially
attributed to the method used for handling missing time-series data.
We utilized linear imputation to estimate missing values for the number
of visited locations and distance traveled. Because depression often
presents with sudden behavioral anomalies--such as abrupt social
withdrawal or periods of prolonged immobility--linear imputation
may have inadvertently smoothed out these critical, non-linear behavioral
spikes. Future research utilizing digital phenotyping should explore
non-linear or behaviorally-aware imputation techniques to better preserve
the natural variance and anomalies inherent in psychiatric mobility
data.

\section{Declarations}

\subsection*{Ethical Approval}

This study was approved by the University of New South Wales Human
Research Ethics Committee (Approval No: HC200466). ``Informed consent''
was obtained from all the participants and all methods were carried
out in accordance with relevant guidelines and regulations.

\subsection*{Funding}

This work was funded in part by a grant from the UK Wellcome Trust
(grant number: 303030/Z/23/Z).

\subsection*{Availability of data and materials}

To request access to de-identified data, please contact Professor
Jill Newby, via j.newby@unsw.edu.au.

\section*{Acknowledgment}

The Vibe Up Trial, from which the data analysed in this study were
obtained, was funded by the Medical Research Future Fund {[}MRFAI000028{]}.
The current research was partially supported by the Wellcome Trust
{[}303030/Z/23/Z{]}. AW was funded by a National Health and Medical
Research Council Investigator Grant {[}2017521{]}.

\bibliographystyle{plain}
\bibliography{reference}

@Article{Nguyen2021,
  author    = {Nguyen, Dang and Gupta, Sunil and Rana, Santu and Shilton, Alistair and Venkatesh, Svetha},
  journal   = {Information Sciences},
  title     = {{Fairness Improvement for Black-box Classifiers with Gaussian Process}},
  year      = {2021},
  pages     = {542--556},
  volume    = {576},
  publisher = {Elsevier},
}

@InProceedings{Nguyen2020,
  author    = {Nguyen, Dang and Gupta, Sunil and Rana, Santu and Shilton, Alistair and Venkatesh, Svetha},
  booktitle = {AAAI Conference on Artificial Intelligence (AAAI)},
  title     = {Bayesian optimization for categorical and category-specific continuous inputs},
  year      = {2020},
  number    = {04},
  pages     = {5256--5263},
  volume    = {34},
}

@InProceedings{Xu2019,
  author    = {Xu, Lei and Skoularidou, Maria and Cuesta-Infante, Alfredo and Veeramachaneni, Kalyan},
  booktitle = {Advances in Neural Information Processing Systems (NeurIPS)},
  title     = {{Modeling tabular data using Conditional GAN}},
  year      = {2019},
  volume    = {32},
}

@Article{chawla2002smote,
  author  = {Chawla, Nitesh and Bowyer, Kevin and Hall, Lawrence and Kegelmeyer, Philip},
  journal = {Journal of Artificial Intelligence Research},
  title   = {{SMOTE: synthetic minority over-sampling technique}},
  year    = {2002},
  pages   = {321--357},
  volume  = {16},
}

@InProceedings{he2008adasyn,
  author       = {He, Haibo and Bai, Yang and Garcia, Edwardo and Li, Shutao},
  booktitle    = {IEEE International Joint Conference on Neural Networks (IJCNN)},
  title        = {{ADASYN: Adaptive synthetic sampling approach for imbalanced learning}},
  year         = {2008},
  organization = {IEEE},
  pages        = {1322--1328},
}

@Article{batista2004study,
  author    = {Batista, Gustavo and Prati, Ronaldo and Monard, Maria Carolina},
  journal   = {ACM SIGKDD Explorations Newsletter},
  title     = {A study of the behavior of several methods for balancing machine learning training data},
  year      = {2004},
  number    = {1},
  pages     = {20--29},
  volume    = {6},
  publisher = {ACM},
}

@InProceedings{han2005borderline,
  author       = {Han, Hui and Wang, Wen-Yuan and Mao, Bing-Huan},
  booktitle    = {International Conference on Intelligent Computing},
  title        = {{Borderline-SMOTE: a new over-sampling method in imbalanced data sets learning}},
  year         = {2005},
  organization = {Springer},
  pages        = {878--887},
}

@Article{nguyen2011borderline,
  author    = {Nguyen, Hien and Cooper, Eric and Kamei, Katsuari},
  journal   = {International Journal of Knowledge Engineering and Soft Data Paradigms},
  title     = {Borderline over-sampling for imbalanced data classification},
  year      = {2011},
  number    = {1},
  pages     = {4--21},
  volume    = {3},
  publisher = {Inderscience Publishers},
}

@Article{sauglam2022novel,
  author    = {Sa{\u{g}}lam, Fatih and Cengiz, Mehmet Ali},
  journal   = {Expert Systems with Applications},
  title     = {{A novel SMOTE-based resampling technique trough noise detection and the boosting procedure}},
  year      = {2022},
  pages     = {117023},
  volume    = {200},
  publisher = {Elsevier},
}

@Article{schwan2021perceptions,
  author    = {Schwan, Anna},
  journal   = {The Clearing House: A Journal of Educational Strategies, Issues and Ideas},
  title     = {Perceptions of student motivation and amotivation},
  year      = {2021},
  number    = {2},
  pages     = {76--82},
  volume    = {94},
  publisher = {Taylor \& Francis},
}

@Article{habehh2021machine,
  author    = {Habehh, Hafsa and Gohel, Suril},
  journal   = {Current Genomics},
  title     = {Machine learning in healthcare},
  year      = {2021},
  number    = {4},
  pages     = {291--300},
  volume    = {22},
  publisher = {Bentham Science Publishers direct},
}

@Article{alanazi2022using,
  author    = {Alanazi, Abdullah},
  journal   = {Informatics in Medicine Unlocked},
  title     = {Using machine learning for healthcare challenges and opportunities},
  year      = {2022},
  pages     = {100924},
  volume    = {30},
  publisher = {Elsevier},
}

@Article{fernandez2018smote,
  author  = {Fern{\'a}ndez, Alberto and Garcia, Salvador and Herrera, Francisco and Chawla, Nitesh},
  journal = {Journal of Artificial Intelligence Research},
  title   = {{SMOTE for learning from imbalanced data: progress and challenges, marking the 15-year anniversary}},
  year    = {2018},
  pages   = {863--905},
  volume  = {61},
}

@Article{shvetcov2024passive,
  author    = {Shvetcov, Artur and Funke Kupper, Joost and Zheng, Wu-Yi and Slade, Aimy and Han, Jin and Whitton, Alexis and Spoelma, Michael and Hoon, Leonard and Mouzakis, Kon and Vasa, Rajesh and others},
  journal   = {Frontiers in Psychiatry},
  title     = {Passive sensing data predicts stress in university students: a supervised machine learning method for digital phenotyping},
  year      = {2024},
  pages     = {1422027},
  volume    = {15},
  publisher = {Frontiers},
}

@Article{barnett2018relapse,
  author    = {Barnett, Ian and Torous, John and Staples, Patrick and Sandoval, Luis and Keshavan, Matcheri and Onnela, Jukka-Pekka},
  journal   = {Neuropsychopharmacology},
  title     = {Relapse prediction in schizophrenia through digital phenotyping: a pilot study},
  year      = {2018},
  number    = {8},
  pages     = {1660--1666},
  volume    = {43},
  publisher = {Springer},
}

@Article{jongs2020framework,
  author    = {Jongs, Niels and Jagesar, Raj and van Haren, Neeltje and Penninx, Brenda and Reus, Lianne and Visser, Pieter and van der Wee, Nic and Koning, Ina and Arango, Celso and Sommer, Iris and others},
  journal   = {Translational Psychiatry},
  title     = {A framework for assessing neuropsychiatric phenotypes by using smartphone-based location data},
  year      = {2020},
  number    = {1},
  pages     = {211},
  volume    = {10},
  publisher = {Nature},
}

@Article{muller2021depression,
  author    = {M{\"u}ller, Sandrine and Chen, Xi and Peters, Heinrich and Chaintreau, Augustin and Matz, Sandra},
  journal   = {Scientific Reports},
  title     = {{Depression predictions from GPS-based mobility do not generalize well to large demographically heterogeneous samples}},
  year      = {2021},
  number    = {1},
  pages     = {14007},
  volume    = {11},
  publisher = {Nature},
}

@Article{huckvale2023protocol,
  author    = {Huckvale, Kit and Hoon, Leonard and Stech, Eileen and Newby, Jill and Zheng, Wu Yi and Han, Jin and Vasa, Rajesh and Gupta, Sunil and Barnett, Scott and Senadeera, Manisha and others},
  journal   = {BMJ Open},
  title     = {{Protocol for a bandit-based response adaptive trial to evaluate the effectiveness of brief self-guided digital interventions for reducing psychological distress in university students: the Vibe Up study}},
  year      = {2023},
  number    = {4},
  pages     = {e066249},
  volume    = {13},
  publisher = {British Medical Journal Publishing Group},
}

@Article{raugh2020geolocation,
  author    = {Raugh, Ian and James, Sydney and Gonzalez, Cristina and Chapman, Hannah and Cohen, Alex and Kirkpatrick, Brian and Strauss, Gregory},
  journal   = {Schizophrenia Bulletin},
  title     = {Geolocation as a digital phenotyping measure of negative symptoms and functional outcome},
  year      = {2020},
  number    = {6},
  pages     = {1596--1607},
  volume    = {46},
  publisher = {Oxford University Press},
}

@Article{palmius2016detecting,
  author    = {Palmius, Niclas and Tsanas, Athanasios and Saunders, Kate EA and Bilderbeck, Amy C and Geddes, John R and Goodwin, Guy M and De Vos, Maarten},
  journal   = {IEEE Transactions on Biomedical Engineering},
  title     = {Detecting bipolar depression from geographic location data},
  year      = {2016},
  number    = {8},
  pages     = {1761--1771},
  volume    = {64},
  publisher = {IEEE},
}

@Article{kazijevs2023deep,
  author    = {Kazijevs, Maksims and Samad, Manar},
  journal   = {Journal of Biomedical Informatics},
  title     = {Deep imputation of missing values in time series health data: A review with benchmarking},
  year      = {2023},
  pages     = {104440},
  volume    = {144},
  publisher = {Elsevier},
}

@Article{batista2003balancing,
  author  = {Batista, Gustavo and Bazzan, Ana and Monard, Maria Carolina and others},
  journal = {WoB},
  title   = {Balancing training data for automated annotation of keywords: a case study},
  year    = {2003},
  pages   = {10--8},
  volume  = {3},
}

@Article{misiuk2024multivariate,
  author    = {Misiuk, Benjamin and Tan, Yan Liang and Li, Michael and Trappenberg, Thomas and Alleosfour, Ahmadreza and Church, Ian and Ferrini, Vicki and Brown, Craig},
  journal   = {Marine Geology},
  title     = {{Multivariate mapping of seabed grain size parameters in the Bay of Fundy using convolutional neural networks}},
  year      = {2024},
  pages     = {107299},
  volume    = {472},
  publisher = {Elsevier},
}

@InProceedings{rasmussen2003gaussian,
  author    = {Rasmussen, Carl},
  booktitle = {Summer School on Machine Learning},
  title     = {Gaussian processes in machine learning},
  year      = {2003},
  pages     = {63--71},
}

@Article{shahriari2016taking,
  author  = {Shahriari, Bobak and Swersky, Kevin and Wang, Ziyu and Adams, Ryan and Freitas, Nando},
  journal = {Proceedings of the IEEE},
  title   = {Taking the human out of the loop: A review of bayesian optimization},
  year    = {2016},
  number  = {1},
  pages   = {148--175},
  volume  = {104},
}

@Article{borisov2022deep,
  author    = {Borisov, Vadim and Leemann, Tobias and Se{\ss}ler, Kathrin and Haug, Johannes and Pawelczyk, Martin and Kasneci, Gjergji},
  journal   = {IEEE Transactions on Neural Networks and Learning Systems},
  title     = {Deep neural networks and tabular data: A survey},
  year      = {2022},
  number    = {6},
  pages     = {7499--7519},
  volume    = {35},
  publisher = {IEEE},
}

@Article{whitton2023distinct,
  author  = {Whitton, Alexis and Kumar, Poornima and Treadway, Michael and Rutherford, Ashleigh and Ironside, Manon and Foti, Dan and Fitzmaurice, Garrett and Du, Fei and Pizzagalli, Diego},
  journal = {Molecular Psychiatry},
  title   = {Distinct profiles of anhedonia and reward processing and their prospective associations with quality of life among individuals with mood disorders},
  year    = {2023},
  number  = {12},
  pages   = {5272--5281},
  volume  = {28},
}

@Article{nguyen2025large,
  author  = {Nguyen, Dang and Gupta, Sunil and Do, Kien and Nguyen, Thin and Braund, Taylor and Whitton, Alexis and Venkatesh, Svetha},
  journal = {arXiv preprint arXiv:2510.09783},
  title   = {Large Language Models for Imbalanced Classification: Diversity makes the difference},
  year    = {2025},
}

@InProceedings{Yang2024,
  author    = {Yang, June Yong and Park, Geondo and Kim, Joowon and Jang, Hyeongwon and Yang, Eunho},
  booktitle = {ICLR},
  title     = {Language-Interfaced Tabular Oversampling via Progressive Imputation and Self-Authentication},
  year      = {2024},
}

@Article{newby2025brief,
  author  = {Newby, Jill and Gupta, Sunil and Hoon, Leonard and Zheng, WuYi and Whitton, Alexis and Huckvale, Kit and Stech, Eileen and Mackinnon, Andrew and Senadeera, Manisha and Shvetcov, Artur and others},
  journal = {JAMA Network Open},
  title   = {{Brief Digital Interventions for Psychological Distress: An AI-Enhanced Response-Adaptive Randomized Clinical Trial}},
  year    = {2025},
  number  = {10},
  pages   = {e2540502--e2540502},
  volume  = {8},
}

@Article{curtiss2024optimizing,
  author  = {Curtiss, Joshua and Smoller, Jordan and Pedrelli, Paola},
  journal = {Psychological Medicine},
  title   = {Optimizing precision medicine for second-step depression treatment: A machine learning approach},
  year    = {2024},
  number  = {10},
  pages   = {2361--2368},
  volume  = {54},
}

@Article{shamshuzzoha2025novel,
  author  = {Shamshuzzoha, Md and Audry, Tazkia Tasnim Bahar and Alam, Md Jahangir and Bhuiyan, Zaheed Ahmed and Islam, Md Motaharul and Hassan, Mohammad Mehedi},
  journal = {Acta Psychologica},
  title   = {{A novel framework for seasonal affective disorder detection: Comprehensive machine learning analysis using multimodal social media data and SMOTE}},
  year    = {2025},
  pages   = {105005},
  volume  = {256},
}

@Article{zhou2021machine,
  author  = {Zhou, Shuzhe and Ma, Qinhong and Lou, Yiwei and Lv, Xiaozhen and Tian, Hongjun and Wei, Jing and Zhang, Kerang and Zhu, Gang and Chen, Qiaoling and Si, Tianmei and others},
  journal = {Journal of Affective Disorders},
  title   = {Machine learning to predict clinical remission in depressed patients after acute phase selective serotonin reuptake inhibitor treatment},
  year    = {2021},
  pages   = {372--379},
  volume  = {287},
}

@Article{benoit2022using,
  author  = {Benoit, James and Dursun, Serdar and Greiner, Russell and Cao, Bo and Brown, Matthew and Lam, Raymond and Greenshaw, Andrew},
  journal = {The Canadian Journal of Psychiatry},
  title   = {Using machine learning to predict remission in patients with major depressive disorder treated with desvenlafaxine},
  year    = {2022},
  number  = {1},
  pages   = {39--47},
  volume  = {67},
}

@Article{wang2024examining,
  author  = {Wang, Junying and Wu, David and DeLorenzo, Christine and Yang, Jie},
  journal = {Plos One},
  title   = {Examining factors related to low performance of predicting remission in participants with major depressive disorder using neuroimaging data and other clinical features},
  year    = {2024},
  number  = {3},
  pages   = {e0299625},
  volume  = {19},
}

@Article{park2026prediction,
  author  = {Park, Jin-Hyun and Kang, Hee-Ju and Jeon, Ji Hyeon and Kang, Sung-Gil and Kim, Ju-Wan and Kim, Jae-Min and Lee, Hwamin},
  journal = {JMIR Mental Health},
  title   = {Prediction of 12-Week Remission in Patients With Depressive Disorder Using Reasoning-Based Large Language Models: Model Development and Validation Study},
  year    = {2026},
  number  = {1},
  pages   = {e83352},
  volume  = {13},
}

@Article{kautzky2021combining,
  author  = {Kautzky, Alexander and M{\"o}ller, Hans-Juergen and Dold, Markus and Bartova, Lucie and Seem{\"u}ller, Florian and Laux, Gerd and Riedel, Michael and Gaebel, Wolfgang and Kasper, Siegfried},
  journal = {Acta Psychiatrica Scandinavica},
  title   = {Combining machine learning algorithms for prediction of antidepressant treatment response},
  year    = {2021},
  number  = {1},
  pages   = {36--49},
  volume  = {143},
}

@Article{carr2025optimizing,
  author  = {Carr, Ewan and Rietschel, Marcella and Mors, Ole and Henigsberg, Neven and Aitchison, Katherine and Maier, Wolfgang and Uher, Rudolf and Farmer, Anne and Mcguffin, Peter and Iniesta, Raquel},
  journal = {American Journal of Medical Genetics Part B: Neuropsychiatric Genetics},
  title   = {Optimizing the prediction of depression remission: a longitudinal machine learning approach},
  year    = {2025},
  number  = {3},
  pages   = {e33014},
  volume  = {198},
}

@Article{calderon2026baseline,
  author  = {Calderon, Adam and Zainal, Nur Hani and Lu, Chenyang and Fitzsimmons-Craft, Ellen and Wilfley, Denise and Eisenberg, Daniel and Taylor, Barr and Newman, Michelle},
  journal = {OSF},
  title   = {Baseline machine learning prediction of 2-year remission from anxiety, depression, and eating disorders among college students after population-based guided self-help: A secondary analysis of a randomized controlled trial},
}

@Article{manger2019lifestyle,
  author  = {Manger, Sam},
  journal = {Australian Journal of General Practice},
  title   = {Lifestyle interventions for mental health},
  year    = {2019},
  number  = {10},
  pages   = {670--673},
  volume  = {48},
}

@Article{Blumenthal1999,
  author  = {Blumenthal, James and Babyak, Michael and Moore, Kathleen and Craighead, Edward and Herman, Steve and Khatri, Parinda and Waugh, Robert and Napolitano, Melissa and Forman, Leslie and Appelbaum, Mark and others},
  journal = {JAMA Internal Medicine},
  title   = {Effects of exercise training on older patients with major depression},
  year    = {1999},
  number  = {19},
  pages   = {2349--2356},
  volume  = {159},
}

@Article{eisendrath2016randomized,
  author  = {Eisendrath, Stuart and Gillung, Erin and Delucchi, Kevin and Segal, Zindel and Nelson, Craig and McInnes, Alison and Mathalon, Daniel and Feldman, Mitchell},
  journal = {Psychotherapy and Psychosomatics},
  title   = {A randomized controlled trial of mindfulness-based cognitive therapy for treatment-resistant depression},
  year    = {2016},
  number  = {2},
  pages   = {99--110},
  volume  = {85},
}

@Article{menardi2014training,
  author  = {Menardi, Giovanna and Torelli, Nicola},
  journal = {Data Mining and Knowledge Discovery},
  title   = {Training and assessing classification rules with imbalanced data},
  year    = {2014},
  number  = {1},
  pages   = {92--122},
  volume  = {28},
}

@InProceedings{wilson2016deep,
  author    = {Wilson, Andrew Gordon and Hu, Zhiting and Salakhutdinov, Ruslan and Xing, Eric},
  booktitle = {AISTAT},
  title     = {Deep kernel learning},
  year      = {2016},
  pages     = {370--378},
}

@InProceedings{snelson2005sparse,
  author    = {Snelson, Edward and Ghahramani, Zoubin},
  booktitle = {NeurIPS},
  title     = {{Sparse Gaussian processes using pseudo-inputs}},
  year      = {2005},
  volume    = {18},
}

@InProceedings{wipf2007new,
  author    = {Wipf, David and Nagarajan, Srikantan},
  booktitle = {NeurIPS},
  title     = {A new view of automatic relevance determination},
  year      = {2007},
  volume    = {20},
}

@Article{rush2006acute,
  author    = {Rush, John and Trivedi, Madhukar and Wisniewski, Stephen and Nierenberg, Andrew and Stewart, Jonathan and Warden, Diane and Niederehe, George and Thase, Michael and Lavori, Philip and Lebowitz, Barry and others},
  journal   = {American Journal of Psychiatry},
  title     = {{Acute and longer-term outcomes in depressed outpatients requiring one or several treatment steps: a STAR* D report}},
  year      = {2006},
  number    = {11},
  pages     = {1905--1917},
  volume    = {163},
  publisher = {American Psychiatric Association},
}

@Article{lovibond1995manual,
  author  = {Lovibond, Sydney},
  journal = {Sydney Psychology Foundation},
  title   = {Manual for the depression anxiety stress scales},
  year    = {1995},
}

\end{document}